\UseRawInputEncoding
\pdfoutput=1

\documentclass[journal]{IEEEtran}
\usepackage[T1]{fontenc}
\usepackage{times}
\usepackage{epsfig}
\usepackage{graphicx}
\usepackage{amsmath}
\usepackage{amssymb}
\usepackage{subfigure}
\usepackage{multirow}
\usepackage{booktabs}
\ifCLASSINFOpdf
\else
\fi
\begin{document}
%
\title{Contrastive Cross-Modal Knowledge Sharing Pre-training 
for Vision-Language \\Representation Learning and Retrieval}


\author{\IEEEauthorblockN{
Keyu Wen, Zhenshan Tan, Qingrong Cheng, Cheng Chen, and Xiaodong Gu}

\thanks{The authors are with Department of Electronic Engineering,
Fudan University, Shanghai 200438, China  (email: kywen19, zstan19, qrcheng17, chengchen19@fudan.edu.cn, xdgu@fudan.edu.cn). Corresponding author: Xiaodong Gu.}
}


%



\IEEEtitleabstractindextext{%
\begin{abstract}
Recently, the cross-modal pre-training task has been a hotspot because of its wide application in various down-streaming researches including retrieval, captioning, question answering and so on. However, exiting methods adopt a one-stream pre-training model to explore the united vision-language representation for conducting cross-modal retrieval, which easily suffer from the calculation explosion. Moreover, although the conventional double-stream structures are quite efficient, they still lack the vital cross-modal interactions, resulting in low performances.
Motivated by these challenges, we put forward a Contrastive Cross-Modal Knowledge Sharing Pre-training (COOKIE) to grasp the joint text-image representations. Structurally, COOKIE adopts the traditional double-stream structure because of the acceptable time consumption. To overcome the inherent defects of double-stream structure as mentioned above, we elaborately design two effective modules. Concretely, the first module is a weight-sharing transformer that builds on the head of the visual and textual encoders, aiming to semantically align text and image. This design enables visual and textual paths focus on the same semantics. The other one is three specially designed contrastive learning, aiming to share knowledge between different models. The shared cross-modal knowledge develops the study of unimodal representation greatly, promoting the single-modal retrieval tasks. Extensive experimental results on multi-modal matching researches that includes cross-modal retrieval, text matching, and image retrieval reveal the superiors in calculation efficiency and statistical indicators of our pre-training model. The fine-tuned COOKIE on three prevailing cross-modal datasets achieves better performances. Especially, compared with the mainstream one-stream models, we only use 3/1000 inference time. Besides, considering image retrieval and text matching tasks for single-modal matching, the percentage gains also reach 5.7\% and 3.9\%.
\end{abstract}

}

\maketitle

\IEEEdisplaynontitleabstractindextext

%
\IEEEpeerreviewmaketitle

\section{Introduction}

\IEEEPARstart{C}{ross-modal} pre-training aims to narrow the diverse gap between vision and texts \cite{unicoder,uniter,oscar}. Benefitting from this favorable characteristic, it significantly promotes the development of representation learning in the field of vision-language. Recently, the large-scale image-text pairs are utilized by vision-language pre-training (VLP) models to explore the common representation of visual input and textual input, which benefit down-streaming V+L tasks, such as cross-modal retrieval \cite{mhtn,sman,yu2020reasoning,qrc,zhang2019multi}, visual question answering \cite{vqa1,vqa2} and image captioning \cite{image-caption1,image-caption2}. In this paper, we propose a pre-training method for one of the most important down-streaming tasks, \textit{i.e., } retrieval task, including image-text matching and video-text matching of cross-modal retrieval and text matching and image retrieval of single-modal matching.


\begin{figure}[t]
\centering
\includegraphics[width=1\linewidth]{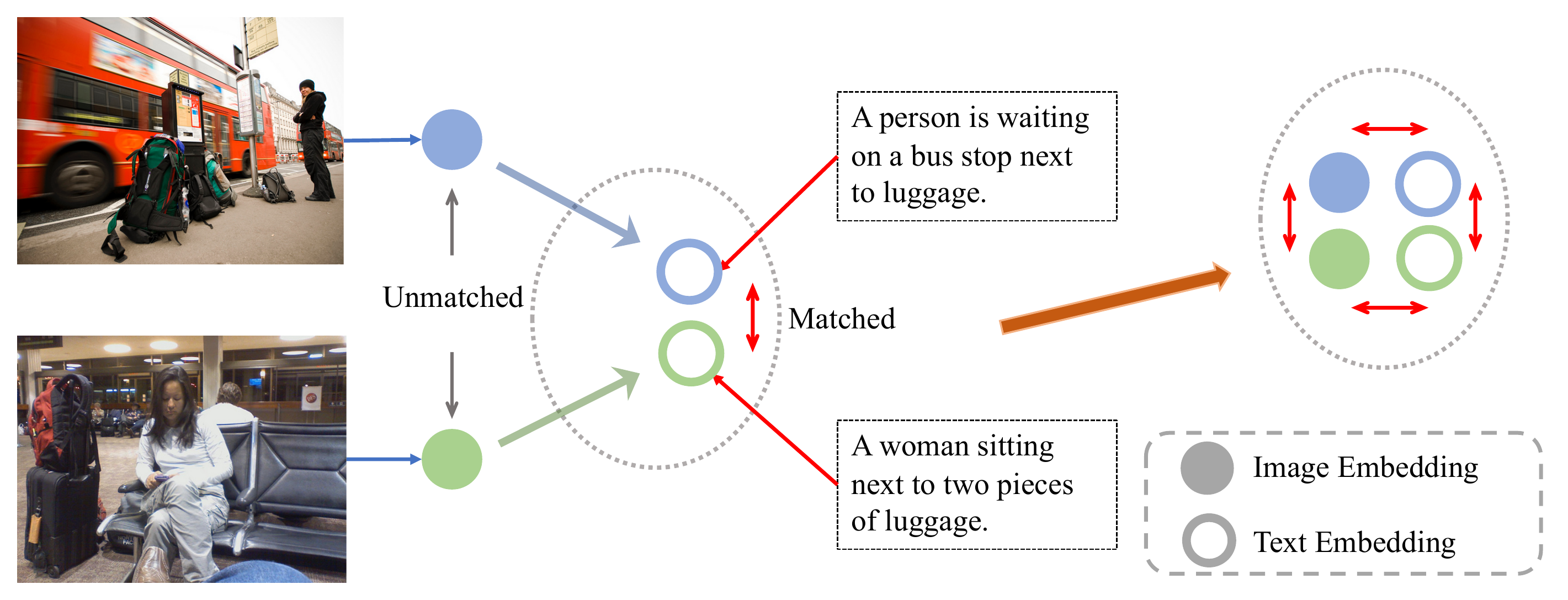}
\caption{Illustration of the proposed cross-modal knowledge sharing. Generally, semantically similar images sometimes differ in structure, texture, style and background, leading to inaccurate matching. Cross-modal knowledge sharing can narrow the embedding distance in common space of corresponding texts by matching the semantics.}\label{intro}
\end{figure}


\begin{figure*}[htbp]
\centering
\includegraphics[width=1\linewidth]{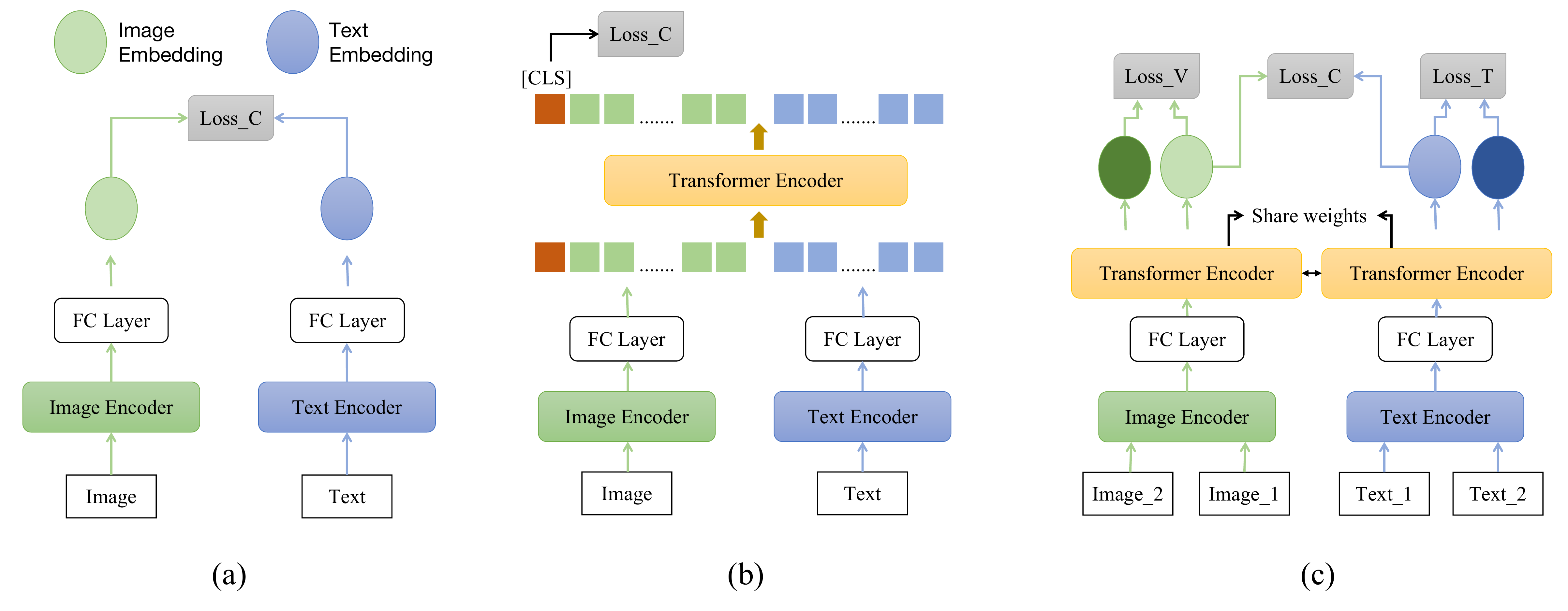}
\caption{Various structures to conduct the vision-language pre-training. $Loss\_C$ denotes cross-modal loss while $Loss\_V$ and $Loss\_T$ are unimodal visual and textual loss.  (a) Double-stream structure. Images and texts in the structure are separately fed into the common subspace, which computes the similarity with a metric function such as cosine similarity.  (b) One-stream structure. A unified transformer encoder in the structure encodes the concatenation of visual token and textual token and then provides a learned similarity score.  (c) The proposed COOKIE. COOKIE is based on a double-stream structure, which implicitly achieves the cross-modal interaction by adding a weight-sharing transformer encoder. Furthermore, we employ three kinds of losses, including within-modal losses, to promote representation learning collaboratively.}\label{structure}
\end{figure*}

We human beings usually perceive the real-world with multiple senses. Similarly, the model pre-training should use multi-model supervision. As shown in Fig. ~\ref{intro}, although the two images share the same semantics in our perspective, they look completely different. Therefore, we should resort to cross-modal pre-training instead of one-modal pre-training. In recent works, the one-steam VLP models are applied for cross-modal pre-training. As illustrated in Fig. ~\ref{structure}, they adopt multi-layer transformer encoder \cite{transformer} and concatenate the inputs of visual and texture tokens. However, these methods still suffer from two obvious issues: a) For two-stage feature extraction, previous method \cite{dsran} has pointed out that the model with Faster R-CNN \cite{faster} will consume extra inference time and lose the vital global information. b) For one-stream methods, the concatenation of image token and text token of retrieval tasks leads to inference time explosion. Also, some researchers apply double-stream methods to the cross-modal pre-training. As seen in Fig. ~\ref{structure} (a), they utilize the visual and textual paths to encode the images and texts, respectively. However, despite the soundness in high efficiency, their performances in cross-modal retrieval are limited. They have two constraints: a) The semantic alignments of image and text will be weakened by the model without cross-modal interactions. b) The vital information of the single-modal encoder learning from the original image or text may be lost by the simple supervision of cross-modal contrastive learning (CCL).




In this paper, we put forward a \textbf{C}ontrastive Cr\textbf{o}ss-M\textbf{o}dal \textbf{K}nowledge Shar\textbf{i}ng Pr\textbf{e}-training (COOKIE) framework for multi-modal retrieval tasks. The proposed COOKIE takes advantage of both the double-stream model and the one-stream VLP model while avoiding their disadvantages mentioned above. We elaborately design two effective parts in COOKIE, including a double-stream visual semantic embedding structure with a weight-sharing transformer encoder (WS-TE) and both cross-modal and single-modal contrastive learning methods. Fig.~\ref{structure} visualizes the differences between our structures and others. 


The first part aims to speed up the training and inference and strengthen the semantic alignment of image and text. Concretely, our framework is based on the double-stream manner, which avoids the inference time explosion that usually occurs in the one-stream method. For visual stream, COOKIE extracts feature with ResNet instead of Faster-RCNN, which avoids the expensive computation cost and keeps the global visual information. Considering the deficiency of cross-modal interactions of previous one-stream methods, we design a WS-TE to make the model focus on the tokens with similar semantics, aiming to guarantee the refined vision-language alignment.


The second part is the contrastive learning designs, including a cross-modal contrastive learning (CCL), a single-modal visual contrastive learning (VCL), and a textual contrastive learning (TCL). Different from previous single-modal methods \cite{csq,bert,clear}, the knowledge of the pre-trained image encoder (ResNet) and text encoder (BERT) in our cross-modal contrastive pre-training is shared. Fig.~\ref{intro} provides the explanation of cross-modal knowledge sharing, the two images with the same semantics, \textit{ ``a person is waiting with luggage''}. However, they present different performances in different images, such as camera angle and background. With the aid of CCL, the image features are pushed close with the lead of text embeddings. Besides, for single-modal encoders, the information explored from large-scale unimodal pre-training is also important. Therefore, we add VCL and TCL to maintain the important single-modal knowledge of the original image and text. The proposed single-modal objectives are different from structure-preserving losses \cite{spl1,spl2}. They improve the alignment of cross-modal semantics by manually searching for the positive within-modal pairs. Differently, the automatically generated pairs of our method are simpler and more effective. Furthermore, the proposed VCL and TCL are able to explore within-modal similarity by the visual and textual encoder, which benefits single-modal retrieval tasks.

Compared with our previous work \cite{cookie}, this paper illustrates more details of the pre-training structure and explains the advantages over previous structures. At the same time, we analyze the used pre-training corpus. For the experiment part \ref{experiment}, more systematic discussions and results are given to demonstrate the effectiveness of the proposed structure and the pre-training mode.



To summarize, the contributions can be concluded as follows.
\begin{itemize}
\item This paper proposes a novel cross-modal pre-training paradigm COOKIE. With the aid of weight-sharing transformer encoder (WS-TE), our method can combine the calculational efficiency of the double-stream structure and the comparative effectiveness of the one-stream structure.
\item For cross-modal knowledge sharing, we design three pre-training objectives including cross-modal contrastive learning (CCL) and single-modal contrastive learning (VCL and TCL), aiming to promote multi-modal retrieval.
\item Our method achieves consistently superior performances on various multi-modal matching tasks, including image-text matching, video-text matching, text matching, and image retrieval. Concretely, the proposed framework COOKIE achieves competitive results compared to the SOTA method Oscar \cite{oscar}, while we only use 3/1000 inference time on two prevailing testing datasets Flickr30K and MSCOCO. Our method increases the indicator $R@1$ of MSRVTT $16.0$ to $20.0$. Besides, the percentage of gains reaches 5.7\% and 3.9\% on image retrieval and text matching of the single-modal matching task, respectively.
\end{itemize}

\begin{figure*}[htbp]
\centering
\includegraphics[width=1\linewidth]{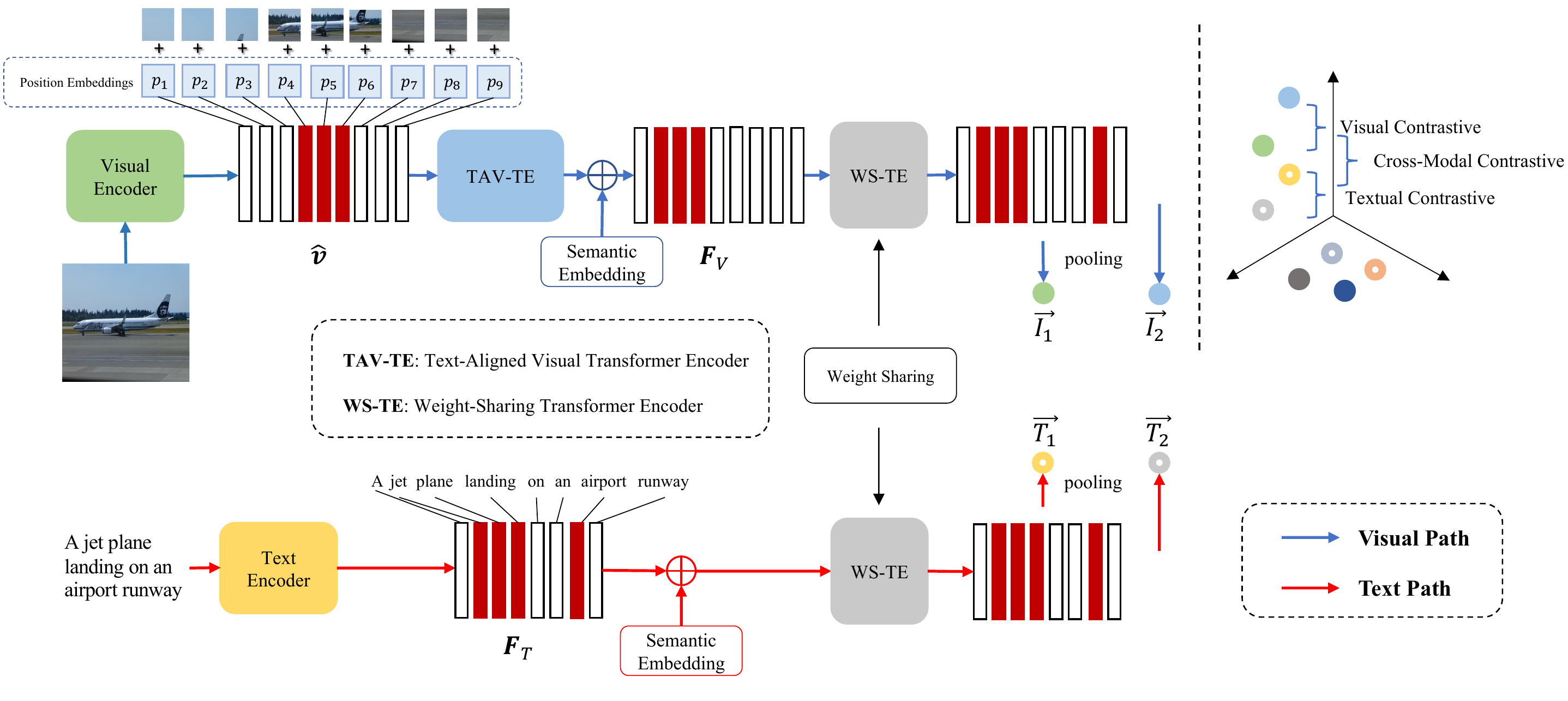}
\caption{The proposed COOKIE contains two paths, the visual and the textual path. For the visual path, firstly a CNN extracts the patch features from the raw image, followed by a text-aligned visual transformer encoder and a weight-sharing transformer. The textual path consists of a BERT encoder and the same weight-sharing transformer. Three contrastive objectives are designed to supervise the pre-training.}\label{model}
\end{figure*}

\section{Related Work}
\subsection{Multi-modal Retrieval and Matching}

The multi-modal matching and retrieval tasks contain cross-modal matching, single-modal retrieval, and matching \cite{cheng2022semantic}. Further, they can be broken down into more sub-directions. This paper discusses four mainstreams, including image-text matching, video-text matching, image retrieval, and text matching. These experiments can verify the efficiency and effectiveness of the proposed framework. Next, we will briefly outline the development of these areas. The basic paradigm of cross-modal retrieval is initially created by CCA \cite{cca}. They feed the images and texts into a common space and then calculate their similarity. Recently, the hinge-based hard triplet loss is adopted by Faghri et al. \cite{vse++} to act as a baseline for most of the recent research. PFAN++ \cite{pfan++} employs a position attention mechanism to learn the representations between image regions and position cells. More recently, the regional visual features attract wide attention. For example, SCAN \cite{scan} introduces the object detection method Faster R-CNN \cite{faster} to exploit the regional visual features, which can extract fine-grained visual-textual matching. In addition, Liu et al. \cite{focus} also propose a focal attention mechanism based on the regional feature to learn more accurately semantic alignment between regional and textual features. Besides, the graph convolutional network (GCN) \cite{gcn} and graph attention network (GAT) \cite{gat} are applied to the retrieval and matching tasks. VSRN \cite{vsrn} and DSRAN \cite{dsran} introduce GCN or GAT to the network to learn the representations within images to better align with texts. Similar to image-text matching, video-text matching also divides the videos into frames and then processes the obtained frames with pooling strategies \cite{hgr,gpo}. For single-modal retrieval and matching, image retrieval needs to find the most relevant images in the image query \cite{dubey2020decade}. Yuan et al. \cite{csq} propose a central similarity quantization for the image retrieval task. Another widely studied task in single-modal retrieval and matching is semantic text similarity (STS), which aims at measuring the similarity between the two given sentences. As did in \cite{sbert}, the similarity score of STS can be offered by either directly obtaining the output of a BERT \cite{bert} or computing the cosine similarity between the two text embeddings with a siamese BERT.

\subsection{Cross-modal Pre-training}

Inspired by previous single-modal pre-training works such as visual contrastive learning \cite{simclr,moco,simsiamese} and textual language modeling \cite{gpt,bert}, we pre-train our cross-modal encoders with large-scale image-text pairs. The pre-trained cross-modal encoders can be decomposed into two kinds of architectures, including the two-stream architecture for image-text independently and the one-stream architecture for both images and texts. They usually take the visual regional features and the word embeddings as the inputs. Some researchers \cite{vilbert,lxmert} extract the representations of image and text with two transformers and then learn the coupling with a unified transformer. While others \cite{oscar,uniter} directly feed the concatenation of regional features and word embeddings to the network, and then processes them with a transformer-based model. With the aid of cross-modal interaction during processing, these methods achieve promising results. However, they still suffer from time explosion during inference. Different from the above methods, we propose another more effective retrieval method. The raw images and sentences are directly used as the inputs, and then they are processed by a weight-sharing transformer in a two-stream manner.

\subsection{Contrastive Learning}

Three kinds of contrastive learning are conducted in this paper, including visual contrastive learning (VCL), textual contrastive learning (TCL), and cross-modal contrastive learning (CCL). Recently, contrastive learning has been widely concerned by researchers because of its excellent performance in representation learning and content-based retrieval. Specifically, VCL aims at minimizing the distance of learned representations between the raw image and augmented image \cite{simclr, moco,infonce,simsiamese}. The classic VCL is proposed by Chen et al. \cite{simclr}. They try to maximize the similarity between an image and its augmented transformation. He et al. \cite{moco} employ a momentum-based contrastive learning method, which constructs a dynamic dictionary to promote unsupervised contrastive learning. For TCL, Wu et al. \cite{clear} prove its effectiveness based on BERT \cite{bert}. Even to the extent that, the simple augmentations with dropout \cite{simcse} also work well in sentence representation learning. As discussed in Section~\ref{3.2}, CCL learns a common subspace of visual modality and textual modality. For instance, Unimo \ cite {unimo} uses CCL to carry out single-mode, multi-mode understanding, and generation tasks concurrently. CCL can also be applied to text-to-image synthesis \cite{xgcgan}. All of these have been elaborately devised for effective and efficient multi-modal retrieval and matching methods.	

\section{The Proposed Method}
%
In this section, we present a comprehensive introduction to the proposed contrastive cross-modal knowledge sharing pre-training for vision-language representation and retrieval. 
Firstly in Section \ref{3.1}, the model architecture is detailed. It contains an image encoder, a text encoder, a text-aligned visual transformer encoder and a weight-sharing transformer encoder. Then in Section \ref{3.2}, we introduce cross-modal contrastive pre-training, which performs important work in aligning cross-modal features and transferring knowledge. Finally in Section \ref{3.3}, single-modal contrastive pre-training including visual and textual contrastive learning is detailed. 

\subsection{Overall Structure}\label{3.1}
The overall method is shown in Fig.~\ref{model}. Previously,  like did in \cite{vilbert, uniter,oscar}, researchers extract visual regional features with Faster R-CNN \cite{faster} and concatenate them with textual word embeddings as the input. Then transformer-based models \cite{bert} are used to process the input. 
For our method,  an original ResNet \cite{resnet, resnext} or BERT \cite{bert} is used to separately encode images and texts. Specifically, having a pair of image and sentence $(V,C)$, the purpose is to get the individual visual or textual embedding $\Vec{I}$ and $\Vec{T}$, which serves as the final representations for cross-modal retrieval as well as single-modal retrieval.





\paragraph{Visual Representation Learning}
We directly extract visual features with a ResNet. Inspired by BUTD(bottom-up and top-down) attention \cite{butd}, previous VLP methods extract visual regional features with a Faster-RCNN pretrained on Visual-Genome dataset \cite{visual-genome}, which makes a two-stage situation for training and inference process. To maintain efficiency, we take an end-to-end approach, which at the same time gets more global features from the raw image. Specifically, the fully-connected layer at the end of ResNet \cite{resnet} or ResNeXt \cite{resnext} is removed and the output feature is flattened without pooling. Then we get the visual patch features $\textbf{v}=\{v_1,v_2,...,v_n\} \in \mathbb{R}^{D_V}$. Here $n$ is the number of patches and $D_V$ is the dimension of the visual feature. We use a fully-connected layer to change the dimension of the features. Additional position embeddings are added to learn the visual features' relative positional relations. The output visual features are $\hat{\textbf{v}}=\{\hat{v_1},\hat{v_2},...,\hat{v_n}\} \in \mathbb{R}^{D}$.


\begin{equation}
    \hat{v_i}=v_i W_V+b_V+p_i,
\end{equation}
where $W_V \in \mathbb{R}^{D_V \times D}, b_V, p_i \in \mathbb{R}^{D}$. $W_V$ and $b_V$ are the parameters of FC layer and $p_i$ is the position embedding for the $i_{th}$ patch.


\paragraph{Textual Representation Learning}
For text features, we use pre-trained BERT-base model \cite{bert} and take the output of the last layer. They are denoted as $\textbf{t}=\{t_1,t_2,...,t_m\} \in \mathbb{R}^{D_T}$, where $m$ is the maximum number of words in one sentence and $D_T$ is the dimension of the word embedding. An FC layer projects the textual features into the common latent space with visual modality. Then we get the textual features denoted as  $\textbf{F}_T=\{f_{T_1},f_{T_2},...,f_{T_m}\} \in \mathbb{R}^{D}$.


\begin{equation}
\label{eqn2}
    f_{T_i}=t_i W_T + b_T + s_T,
\end{equation}
where $W_T \in \mathbb{R}^{D_T \times D}, b_T \in \mathbb{R}^{D}$ in Eq.~\ref{eqn2} are the parameters of FC layer. Following \cite{pixel-bert}, a textual semantic embedding vector $s_T$ is added to the features. 


\paragraph{Text-Aligned Visual Transformer}
For images, the commonly used CNN is a local operator which conducts sliding window on the images. While for texts, the transformer layer of BERT model is a global operator. This leads to textual features from BERT having a different distribution from image features from CNNs. In order to align them, a text-aligned visual transformer encoder (TAV-TE) is added. TAV-TE aims to provide the image side with a global attention calculation. The transformer encoder (TE) in this paper follows the standard definition \cite{transformer}. We add a visual semantic embedding vector $s_V$ to the features.


\begin{equation}
\textbf{F}_V=TE_{TAV} (\hat{\textbf{v}})+s_V.
\end{equation}




\paragraph{Weight-Sharing Transformer}
Aiming at prompting visual and textual paths to focus on the same semantics, a weight-sharing transformer encoder (WS-TE) was added on top of the network. Consisting of a multi-head self-attention process and an FFN(feed-forward network), WS-TE can make the input tokens pay more attention to salient areas. The convolution kernels of CNN share weights, reducing parameters and enabling translation equivariant \cite{cnn-equiv}. In other words, the network extracts the same features no matter how the image translates. Here with the parameter-sharing scheme for image and text tokens, the self-attention layer gives relative attention values for analogous semantics of images and texts. If the same semantics of vision and language modalities are given greater weights, the final representation will also be better aligned, resulting in better aligning visual and textual representations.



\begin{equation}
\vec{I}=Pooling (TE_{WS} (\textbf{F}_V)),
\vec{T}=Pooling (TE_{WS} (\textbf{F}_T)).
\end{equation}

\subsection{Cross-Modal Contrastive Learning}\label{3.2}
The aim of cross-modal contrastive learning is to construct a common subspace where the heterogeneous gap is narrowed for images and texts.


At the same time, such a learning process enables cross-modal knowledge sharing, namely from image to language understanding and vice versa.

The whole network, including the image and text encoders and the weight-shared TE are optimized with info-nce loss \cite{infonce}, a commonly used loss function in contrastive learning. For $L_{i2t}$, the positive sample is the matched text and the negative samples are the remaining texts in the mini batch. For $L_{t2i}$, vice versa.

\begin{equation}
\begin{aligned}
    &L_{infonce} (q,k)=\\
    &-\frac{1}{N}\sum_{i=1}^N log \frac{exp (q\cdot k^+/\tau)}{exp (q\cdot k^+/\tau) + \sum_{j=1}^{N-1}exp (q\cdot k^-/\tau)},
\end{aligned}
\end{equation}

\begin{equation}
    L_{i2t}=L_{infonce} (\Vec{I},\Vec{T}),
\end{equation}
    

\begin{equation}
    L_{t2i}=L_{infonce} (\Vec{T},\Vec{I}).
\end{equation}

Here $N$ is the size of one mini batch. $\tau$ is a temperature hyper-parameter. $'+'$ and $'-'$ refer to the positive sample and the negative sample respectively. 

\subsection{Single Modal Contrastive Learning}\label{3.3}
Cross-modal contrastive learning is a key role for cross-modal retrieval and promotes knowledge sharing between image encoder and text encoder. However, to get single-modal representations used for single-modal tasks like image retrieval and text matching, the uni-modal encoders should not lose too much original information learned from single-modal data. To maintain the single-modal encoder's ability to process its own modal data, we design visual and textual contrastive learning.


\begin{table}[t]\setlength{\tabcolsep}{8.5pt}
\caption{Statistics of the pre-training corpus.} 
        \centering
        
        \fontsize{9.5}{10}\selectfont

        \centering
        \begin{tabular}{l|c|c}
            \toprule
            Dataset  & Image & Text\\
   
            \midrule

            CC (train) & 2.8M & 2.8M \\
            SBU (all) & 0.8M & 0.8M\\
            MSCOCO (train) & 113k & 566k\\
            Flickr30K (train) & 29k & 145k \\
            VQA2.0 (train) & 83k & 444k \\
            GQA (bal-train) & 79k & 1.0M \\
            Total &3.9M&5.9M\\
            \midrule
        \end{tabular}

    \label{tab:pretrain}
    \end{table}

\begin{table}[t]\setlength{\tabcolsep}{3.5pt}
\caption{Experimental settings for all downstream datasets. ITM: image-text matching, VTM: video-text matching, TM: text matching, IR: image retrieval. For ITM and IR, we list num of images. For VTM, we list num of videos. For TM, we list num of text pairs. STS12-16 only contain test set.} 
        \centering
        
        \fontsize{9.5}{10}\selectfont

        \centering
        \begin{tabular}{c|c|c|c|c|c|c}
            \toprule
            Task & Dataset & Train & Test & Retrieval & I:T & Class \\
   
            \midrule
            ITM & MSCOCO & 113,287 & 5,000 & - & 1:5 & -\\
            ITM & Flickr30K & 29,000 & 1,000 & - & 1:5 & -\\
            VTM & MSRVTT & 6,753 & 2,990 & - & 1:20 & -\\
            TM & STSB & 5,749 & 1,379 & - & - & -\\
            IR & MSCOCO & 10,000 & 5,000 & 112,217 & - & 80\\
            IR & NUSWIDE & 10,000 & 2,040 & 149,685 & - & 21\\
            \midrule
        \end{tabular}
        
        
        \centering
        \begin{tabular}{c|c|c|c|c|c}
            \toprule
            Dataset & STS12 & STS13 & STS14 & STS15 & STS16 \\
            \midrule
            Test & 3,108 & 1,500 & 3,750 & 3000 & 1186\\
    
            \midrule
        \end{tabular}
        
        \label{tab:datasets.2}

    \label{tab:datasets}
    \end{table}

\paragraph{Visual Contrastive Learning}
Visual self-supervised learning can effectively promote image recognition \cite{simclr, moco, simsiamese}. We use visual contrastive learning to improve the image encoder's ability to understand images while accepting knowledge from texts. We use two different augmentations of the images as the input. The goal is to draw these two augmentations closer during training. Specifically, the distance between the positive pairs is minimized while the distance of the negative pairs is maximized. Having a raw image $V$, we denote the image encoder together with the weight-sharing TE as $\textbf{E}_V$. The visual info-nce loss is optimized as follows.


\begin{equation}
    \Vec{I_1}=\textbf{E}_V (aug_{v_1} (V)),\Vec{I_2}=\textbf{E}_V (aug_{v_2} (V)),
\end{equation}
\begin{equation}
    L_i = L_{infonce} (\Vec{I_1},\Vec{I_2}),
\end{equation}
where $aug_v (\cdot)$ denotes visual augmentation. In this part, the image augmentation includes random cropping, color jitter, flipping, color dropping, and gaussian blur.
\paragraph{Textual Contrastive Learning}
In the natural language processing field, contrastive learning is not the first choice for self-supervised learning. Instead, the common options are causal language modeling (CLM) \cite{gpt} and masked language modeling (MLM) \cite{bert}. Wu et al. \cite{clear} showed that contrastive learning is effective when it comes to sentence representation. In our pattern, texts are augmented with random masking, substituting, and deleting. These random operations can improve the robustness of the model. Text encoders can capture information from pictures while maintaining attention to the semantic properties of sentences. As with the visual path, we denote the text encoder and weighted sharing TE as $\textbf{E}_T$ and optimize $\textbf{E}_T$ with info-nce loss. Given a raw caption $C$,


\begin{equation}
\label{eqn10}
    \Vec{T_1}=\textbf{E}_T (aug_{t_1} (C)),\Vec{T_2}=\textbf{E}_T (aug_{t_2} (C)),
\end{equation}
\begin{equation}
    L_t = L_{infonce} (\Vec{T_1},\Vec{T_2}),
\end{equation}

Here $aug_t (\cdot)$ in Eq.~\ref{eqn10} denotes text augmentation.

The overall pre-training objective of COOKIE is defined below.

\begin{equation}
    L_{Pre-training}=L_{i2t}+L_{t2i}+L_i+L_t.
\end{equation}

\begin{figure*}[htbp]
\centering\includegraphics[width=1\linewidth]{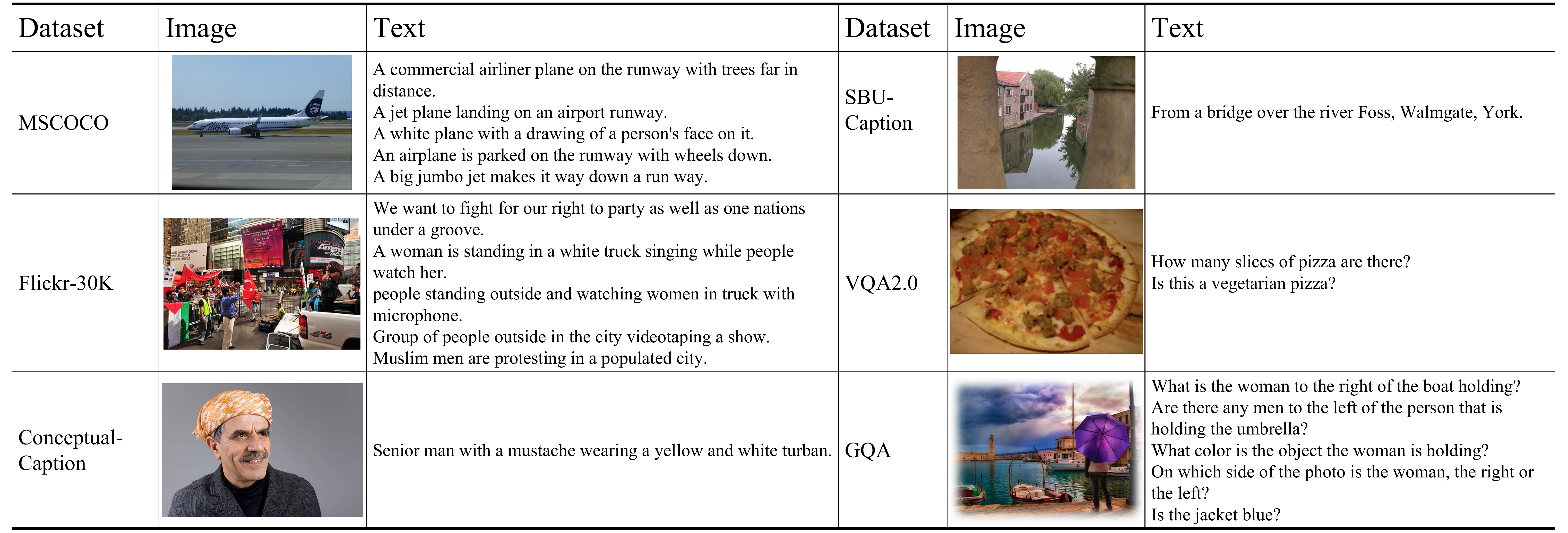}

\caption{Illustration of pre-training datasets. For MSCOCO \cite{mscoco} and Flickr-30K \cite{f30k}, each image has 5 captions. Images in Conceptual-Caption \cite{conceptual} and SBU-Caption \cite{sbu} have one caption each. As for VQA2.0 \cite{vqa2.0} and GQA \cite{gqa}, images are coupled with questions.}\label{datasets}

\end{figure*}

\section{Experiments}\label{experiment}

\subsection{Pre-training Configurations}
\paragraph{Pre-training Corpus}

Several public available image-text datasets Flickr30K \cite{f30k}, Conceptual-Caption(CC) \cite{conceptual}, MSCOCO \cite{mscoco}, SBU-Caption \cite{sbu}, VQA2.0 \cite{vqa2.0} and GQA \cite{gqa} are utilized to pre-train our COOKIE, which yield the total size of 3.9 million images and 5.9 million image-text pairs. We show the details of these datasets in Table~\ref{tab:pretrain}. Noted that, due to broken URLs, for conceptual captions, we only collected 2.8M of the 3M image-text pairs. And for SBU captions, only 0.8M of the 1M pairs are collected. For the top-4 datasets, images are paired with describing captions, which are collected from social media. While for VQA2.0 and GQA, the images are paired with questions about the visual content. Though these questions are raised based on the image, they may contain limited information corresponding to the image. We illustrate these datasets in Fig.~\ref{datasets}.



\paragraph{Implementations}

ResNet50, ResNet101 \cite{resnet} or ResNeXt101 \cite{resnext} are selected as the image encoder and BERT-base \cite{bert} is used as the text encoder. We reshape all images to $512 \times 512$, if not otherwise specified. The dimensions of output features of the image encoder and text encoder $D_V$ and $D_T$ are 2048 and 768, respectively. The dimension of the cross-modal common subspace $D$ is set to 1024. The number of image patches $n$ is set to $16\times 16=256$ and the maximum number of words $m$ is set to 50.  The TAV-TE has 1 layer while the weight-sharing TE has 2 layers. We set the intermediate size and multi-head number to 1024 and 8, respectively.

The model is optimized with AdamW \cite{adamw} for two stages. During the first stage, we merely train the model with $L_{i2t}$ and $L_{t2i}$ for 30 epochs with the batch size set to 576 for stability. In the second stage, the full $L_{Pre-training}$ is used to supervise the training for 10 epochs with the batch size set to 288. The learning rate is 2e-5 initially and decays by 10 times after half of the total epochs for each stage. Noted that, LR for ResNeXt101 is one-tenth of the global LR. We conduct the experiments with Tensorflow v2.2 on 48 Tesla V100 GPUs.



\subsection{Training Details for Downstream Tasks}

In Table~\ref{tab:datasets}, We list the statistics of all the downstream multi-modal matching datasets. Since the number of images and that of texts are unbalanced in ITM and VTM tasks, the column $I:T$ displays the number of captions describing one image. In TM, the evaluation process requires no supervision, so the test sets of STS12-16 and STSB are directly used. For IR, we use the train sets to train the model. Then the test sets serve as the queries retrieving images in the retrieval sets. 


\paragraph{Image-Text Matching}We finetune the pre-trained COOKIE for 20 and 16 epochs on MSCOCO and Flickr30K, respectively. The initial learning rate is 2e-5 or 1e-5 and decays by 10 times after half of the total epochs. AdamW optimizer is used, with the weight decay factor set to 1e-4 and the warm-up proportion set to 0.1. The batch size is set to 384 and 288 for MSCOCO and Flickr30K on ResNeXt-101-based models. We set the batch size to 320 or 240 when it comes to ResNet101-based models. The hinged hard triplet loss \cite{vse++} is defined as follows.


\begin{equation}
\begin{aligned}
L_{imt}=&[\alpha+S (\vec{I^{'}}, \vec{T})-S (\vec{I}, \vec{T})]_+ + \\
&[\alpha+S (\vec{I}, \vec{T}^{'})-S (\vec{I}, \vec{T})]_+, \label{eq_imt} 
\end{aligned}
\end{equation}
where $S (\cdot)$ refers to the similarity function which is cosine similarity in our model. Here $[x]_+\equiv max (x,0)$ and $\alpha$ is the margin, which is set to 0.2. We use $MAX$-$Pooling$ for image-text matching tasks.

\paragraph{Video-Text Matching}
For video-text matching, We adopt the commonly used MSRVTT benchmark and finetune our pre-trained model for 10 epochs. The learning rate is set to 2e-5 which decays by 10 times after 5 epochs. A batch size of 320 is used. The standard split with 6573 videos for training, 2990 for testing, and 497 for validation is adopted. $MEAN$-$Pooling$ or $G$-$Pooling$ is used for video-text matching tasks.


\paragraph{Text Matching}
As sentence embeddings for evaluating semantic text similarity tasks are directly used, a training process is not required. $MAX$-$Pooling$ is adopted for text matching tasks. Pearson and Spearman coefficients are two widely used metrics for evaluating the correlation between the predicted similarity scores and the labels in previous works. Here we use the mean value of them to evaluate.


\paragraph{Image Retrieval}
We use the number of output bits either 16, 32, or 64 for image retrieval. We set the learning rate to 1e-4 initially and it decays with a ratio of 0.7 every 10 epochs. COOKIE is finetuned for a total of 100 epochs with a batch size of 320. $MEAN$-$Pooling$ is used for image retrieval tasks. The task of image retrieval is defined to judge whether the retrieval result is correct according to the category. Thus, we record the common metric $MAP@5000$ for image retrieval.


\begin{table*}[t]\setlength{\tabcolsep}{3pt}
    \fontsize{11}{12}\selectfont
    \centering
    \caption{Results on image-text matching task with MS-COCO $1K$ and $5K$ test set. Here $R101$, $X101$ and $X152$ mean ResNet101, ResNeXt101 and ResNeXt152, respectively. $b$ and $l$ refer to base and large models for Uniter and Oscar. $*$ represents the process of model ensemble. The best results of each metric are marked in bold, while the suboptimal values are underlined.}
    \begin{tabular}{lccccccc|ccccccc}
        \toprule
        \multirow{3}*{\bfseries Methods} &
        \multicolumn{3}{c}{\bfseries Image-to-Text} & \multicolumn{3}{c}{\bfseries Text-to-Image}&
        \multicolumn{1}{c}{\bfseries }&
        \multicolumn{3}{c}{\bfseries Image-to-Text} & \multicolumn{3}{c}{\bfseries Text-to-Image}\\

        \cmidrule{2-15} &
        \multicolumn{7}{c}{\bfseries 1K Test Set} & \multicolumn{7}{c}{\bfseries 5K Test Set}\\

        \cmidrule{2-15} &R@1&R@5&R@10&R@1&R@5&R@10&Rsum&R@1&R@5&R@10&R@1&R@5&R@10&Rsum\\

        \midrule
        \multicolumn{8}{l}{\bfseries Double-Stream Methods} \\
        \midrule
        VSE++ &64.6&90.0&95.7&52.0&84.3&92.0&478.6&41.3&71.1&81.2& 30.3 &59.4&72.4&355.7 \\
        SCAN & 70.9 &94.5 &97.8 &56.4 &87.0 &93.9&500.5& 46.4 &77.4 &87.2 &34.4 &63.7 &75.7&384.0\\
        CVSE & 74.8 &95.1& 98.3 &59.9 &89.4 &95.2&512.7&-&-&-&-&-&-&-\\
        PFAN++$^*$ &77.1&96.5&98.3&62.5&89.9&95.4&519.7&51.2&84.3&89.2&41.4&70.9&79.0&416.0\\
        
        VSRN$^*$ &76.2 &94.8 &98.2 &62.8 &89.7 &95.1&516.8&53.0 &81.1 &89.4 &40.5 &70.6 &81.1&415.7\\
        GSMN$^*$ & 78.4 &96.4& 98.6 &63.3 &90.1 &95.7&522.5&-&-&-&-&-&-&-\\
        DSRAN$^*$ &80.6&96.7&98.7&64.5&90.8&95.8&527.1&57.9&85.3&92.0&41.7&72.7&82.8&432.4\\
        GPO$^{R101}$  & 78.0&95.8&98.5&62.6&90.6&96.0&521.5&56.2&83.7&90.9&40.8&70.6&81.5&423.7 \\
        GPO$^{X101*}$  &85.6 &98.0 &99.4 &73.1 &94.3 &\underline{97.7} &548.1&68.1 &90.2 &95.2 &52.7 &\underline{80.2} &\underline{88.3} &474.8 \\
        \midrule \multicolumn{12}{l}{\bfseries One-Stream Pre-training Methods} \\
        \midrule
        Unicoder-VL &84.3 &97.3 &99.3 &69.7 &93.5 &97.2& 541.3&62.3 &87.1 &92.8 &46.7 &76.0 &85.3&450.2\\
        Pixel-BERT$^{X152}$ &84.9&97.7&99.3&71.6&93.7&97.4&544.6 &63.6&87.5&93.6&50.1&77.6&86.2&458.6\\
        Uniter$^b$ &-&-&-&-&-&-&-&63.3&87.0&93.1&48.4&76.7&85.9&454.4\\
        Uniter$^l$ &-&-&-&-&-&-&-&66.6&89.4&94.3&51.7&78.4&86.9&467.3\\
        Oscar$^b$ &\textbf{88.4}&\textbf{99.1}&\textbf{99.8}&\textbf{75.7}&\textbf{95.2}&\textbf{98.3}&\textbf{556.6} &\underline{70.0}&\textbf{91.1}&\textbf{95.5}&\underline{54.0}&80.0&\textbf{88.5}&\underline{479.1}   \\
        \midrule
        \multicolumn{8}{l}{\bfseries Double-Stream Pre-training Methods} \\
        \midrule
        \textbf{COOKIE$^{R101}$}&81.3&96.2&98.7&67.5&91.5&96.1&531.3 &61.7&86.7&92.3&46.6&75.2&84.1&446.6\\
        \textbf{COOKIE$^{X101}$}&\underline{87.3}&98.1&\underline{99.6}&73.5&94.0&97.5&550.0&69.2&89.6&94.4&52.4&79.6&87.1&472.3\\
        \textbf{COOKIE$^{X101*}$} &\textbf{88.4}&\underline{98.5}&\textbf{99.8}&\underline{75.2}&\underline{94.7}&97.5&\underline{554.1} &\textbf{71.6}&\underline{90.9}&\underline{95.4}&\textbf{54.5}&\textbf{81.0}&88.2&\textbf{481.6}\\

        \midrule
    \end{tabular}
    \centering
    \label{tab:coco}
\end{table*}

\subsection{Data Augmentations for Pre-training}\label{A3}
 Data augmentations for images and texts are adopted in both visual and textual contrastive learning. The augmentation details are shown in the following content.
\paragraph{Visual Augmentations}
We design five data augmentations for visual contrastive learning before resizing the images to a fixed size. All augmentation operations are executed sequentially.

\begin{itemize}
    \item Cropping. we crop each image into a random size $ (\sigma_1 *H,\sigma_2*W)$, where $H$ is the height of the original image, $W$ is the width of original image and $\sigma_1$ and $\sigma_2$ are two scaling factors. For Keeping original information, the two scaling factors are set in the range 0.6-1.
\item Flipping. We randomly flip the images horizontally with 50\% probability.
\item Adding Gaussian noise. We add Gaussian noise to an image With 0.5 probability.
\item Color jitter. Color jitter is performed with a probability of 0.8, which contains changing the brightness, contrast, saturation, and hue of an image.
\item Color dropping. RGB images are converted to gray-scale images with 20\% probability.
\end{itemize}

\paragraph{Textual Augmentations}
Visual augmentation is essential for visual representation learning and it has been successfully applied in many tasks as a common step. However, for texts, classic data augmentations will lead to information loss or significantly change the meaning of the whole sentence. 
Wu et al.\cite{clear} have proved that textual augmentations are also effective in textual contrastive learning. Specifically, each token is processed by textual augmentation with a 20\% probability. The data augmentation procedure of textual contrastive learning is shown as follows.
\begin{itemize}
    \item 50\% of the time: Mask the word with the $[MASK]$ token.
    \item 10\% of the time: Replace the word with a random word chosen from the vocabulary.
    \item 40\% of the time: Directly delete the word. 
\end{itemize}

\subsection{Downstream Matching Tasks}\label{4.2}

\begin{table}[t]\setlength{\tabcolsep}{2pt}
    \caption{Results on image-text matching task with Flickr30K dataset.}
    \fontsize{9.5}{10}\selectfont
    \centering
    \begin{tabular}{lccccccccll}
        \toprule
        \multirow{2}*{\bfseries Methods} &
        \multicolumn{3}{c}{\bfseries Image-to-Text} & \multicolumn{3}{c}{\bfseries Text-to-Image}\\ 
        \cmidrule{2-8} &R@1&R@5&R@10&R@1&R@5&R@10&Rsum \\
        \midrule
        \multicolumn{8}{l}{\bfseries Double-Stream Methods} \\
        \midrule
        VSE++ &52.9 &80.5 &87.2& 39.6 &70.1 &79.5&409.8 \\
        SCAN  &67.9 &89.0 &94.4 &43.9 &74.2 &82.8 &452.2\\
        PFAN++$^*$ &70.1&91.8&96.1&52.7&79.9&87.0&477.6\\
        
        CVSE &73.5 &92.1 &92.1 &52.9 &80.4 &87.8&482.5\\
        VSRN$^*$ &71.3 &90.6 &96.0 &54.7 &81.8 &88.2&482.6\\
        GSMN$^*$ &76.4 &94.3 &97.3 &57.4 &82.3 &89.0&496.8\\
        DSRAN$^*$ &80.5&95.5&97.9&59.2&86.0&91.9&511.0\\
        GPO$^{R101}$ &77.9&93.7&97.4&57.5&83.4&90.2&500.2 \\
        GPO$^{X101*}$ &\underline{88.7} &\textbf{98.9} &\textbf{99.8} &\underline{76.1} &\underline{94.5} &\underline{97.1} &\underline{555.1} \\
        \midrule
        \multicolumn{8}{l}{\bfseries One-Stream Pre-training Methods} \\
        \midrule
        Unicoder-VL &86.2 &96.3&99.0 &71.5 &90.9 &94.9&538.8\\
        Pixel-BERT$^{X152}$ &87.0&\textbf{98.9} &99.5 &71.5 &92.1 &95.8&544.8\\
        Uniter$^b$ &85.9&97.1&98.8&72.5&92.4&96.1&542.8  \\
        Uniter$^l$ &87.3&\underline{98.0}&99.2&75.6&94.1&96.7&550.9 \\
        ERNIE-ViL$^l$ &88.1&98.0&99.2&\textbf{76.7}&93.6&96.4&552.0\\

        \midrule
        \multicolumn{8}{l}{\bfseries Double-Stream Pre-training Methods} \\
        \midrule
        \textbf{COOKIE$^{R101}$} &84.7&96.9&98.3&68.3&91.1&95.2&534.5\\
        \textbf{COOKIE$^{X101}$}&\textbf{89.0}&\textbf{98.9}&\textbf{99.8}&75.6&\underline{94.5}&\underline{97.1}&554.9\\
        \textbf{COOKIE$^{X101*}$}&\textbf{89.0}&\textbf{98.9}&\underline{99.7}&75.6&\textbf{94.6}&\textbf{97.2}&\textbf{555.3}\\
        \midrule

    \end{tabular}
   \label{tab:f30k}
\end{table}

\begin{table}[t]\setlength{\tabcolsep}{2pt}
    \fontsize{9.5}{10}\selectfont
    \centering
    \caption{Results on video-text matching task with MSRVTT dataset. Here $mean$ refers to mean pooling and $gpo$ means the pooling strategy proposed by GPO \cite{gpo}. $*$ means the methods use ResNet-ResNext features and $**$ means the method uses seven-modal features.}
    \begin{tabular}{lccccccccll}
        \toprule
        \multirow{2}*{\bfseries Methods} &
        \multicolumn{3}{c}{\bfseries Video-to-Text} & \multicolumn{3}{c}{\bfseries Text-to-Video}\\ 
        \cmidrule{2-8} &R@1&R@5&R@10&R@1&R@5&R@10&Rsum \\

        \midrule
        MEE$^*$ &13.4&32.0&44.0&6.8&20.7&31.1  & 148.0\\
        VSE++ &14.4&34.1&45.6&8.3&24.0&34.1& 160.5 \\
        HGR &15.0 &36.7 &48.8 &9.2 &26.2 &36.5 &172.4\\
        CE$^{**}$ &15.6&\underline{40.9}&\textbf{55.2}&\underline{10.0}&\underline{29.0}&\textbf{41.2}&\underline{191.9} \\
        GPO &16.0 &38.6& 50.2& 8.7& 25.3 &35.9 &174.7 \\

        W2VV$_{++}^*$ &17.5&40.2&52.5&\textbf{11.1}&\textbf{29.6}&\underline{40.5} &191.4\\
        \textbf{COOKIE} ($mean$) & \underline{19.4}&40.7&51.5&9.8&27.6&38.6&187.6\\
        \textbf{COOKIE} ($gpo$) &\textbf {20.0}&\textbf{42.0}&\underline{54.9}&9.8&28.3&39.6&\textbf{194.6}\\
        \midrule

    \end{tabular}
   \label{tab:msrvtt}
\end{table}

The proposed COOKIE is designed for several multi-modal downstream tasks, including image-text matching, video-text matching, text matching, and content-based image retrieval. All these tasks need high-quality feature representation and fast inference speed. The proposed pre-training strategy can well address these issues. It should be noted that the BERT-base model \cite{bert} is used as the text encoder for all tasks.

\paragraph{Image-Text Matching}
Image-text matching (ITM) requires fine-grained semantic consistency of visual and textual representations. ITM is a fundamental task in visual-textual representation learning, which includes image-to-text retrieval (image as query and text as gallery) and text-to-image retrieval. Following remarkable image-text matching methods \cite{vse++,vsrn,dsran}, a hinged hard triplet loss is chosen to supervise the fine-tuning process. Two widely-used datasets i.e. MSCOCO \cite{mscoco} and Flickr30K \cite{f30k} are chosen to show the performance of ITM. For a fair comparison, the experiments use the same dataset split with previous work \cite{karpathy}. We adopt recall at K ($R@K$) together with $Rsum$ to evaluate the retrieval performance. ResNet101 \cite{resnet} and ResNeXt101 \cite{resnext} are used and marked clearly. We compare the proposed method with double-stream methods without pre-training (VSE++ \cite{vse++}, SCAN \cite{scan}, CVSE \cite{cvse}, PFAN \cite{pfan}, VSRN \cite{vsrn}, GSMN \cite{gsmn}, DARSN\cite{dsran} and GPO \cite{gpo}) and one-stream methods with pre-training (Unicoder-VL \cite{unicoder}, ERNIE-ViL \cite{ernie}, Pixel-BERT \cite{pixel-bert}, Uniter \cite{uniter} and Oscar \cite{oscar}). The overall comparison results are shown in Table \ref{tab:coco} and Table \ref{tab:f30k}. The calculation of $Rsum$ is defined below.

\begin{equation}
\begin{aligned}
Rsum =&\underbrace{R@1 + R@5 + R@10}_{\text{image-to-text retrieval}}+\\&\underbrace{R@1 + R@5 + R@10}_{\text{text-to-image retrieval}}.
\end{aligned}
\end{equation}

\paragraph{Video-Text Matching}
Similar to ITM, video-text matching(VTM) aims at retrieving video by text or vice versa. Compared with ITM, VTM is more challenging and difficult due to the complexity of video data. We conduct experiments on MSRVTT dataset \cite{MSRVTT} to evaluate the performance of the proposed method. For fair comparison, we use ResNet152 \cite{resnet} pre-trained on ImageNet \cite{imagenet} to extract video features by following previous works. Therefore, we do not need an image encoder for VTM. Conditioned on it, the text features are represented by the output of our pre-trained BERT. The final visual and textual representations $\vec{I}$ and $\vec{T}$ are obtained by processing the frame and text features with a pooling strategy. Mean-pooling or g-pooling \cite{gpo} are chosen in our experiments. The model is optimized by the info-nce loss \cite{infonce}. We use the same split with HGR \cite{hgr}, GPO \cite{gpo} for fair comparison. Further, we list some methods using stronger ResNet-ResNeXt features \cite{mee,w2vv} and even seven-modal features \cite{ce}. The comparison results are shown in Table \ref{tab:msrvtt}.



        

   

\begin{table}[t]\setlength{\tabcolsep}{2pt}
    \fontsize{9.5}{10}\selectfont
    \centering
    \caption{Results on STS task. We record the mean values of Pearson and Spearman coefficients.}
        \begin{tabular}{lc|c|c|c|c|c}
        \toprule
         \textbf{Methods}&STS12&STS13&STS14&STS15&STS16&STSB \\

        \midrule
        BERT &28.8&\underline{50.8}&43.9&57.6&58.7&46.1\\

        RoBERTa &47.4 &37.5 &47.9 &55.1 &57.6 &71.9\\
        MACD &-&- &- &- &- &71.8 \\
        CLEAR & \underline{49.0} &48.9 &\underline{57.4} &\underline{63.6} &\underline{65.6}&\underline{72.5}\\
        
        \textbf{COOKIE} & \textbf{63.2}&\textbf{68.0}&\textbf{68.0}&\textbf{72.4}&\textbf{68.1}&\textbf{75.3}
\\
        \midrule

    \end{tabular}
   
   \label{tab:sts}
\end{table}

\begin{table}[t]\setlength{\tabcolsep}{2pt}
    \fontsize{9.5}{10}\selectfont
    \centering
    \caption{Results on image retrieval task with MSCOCO and NUSWIDE datasets.}
        \begin{tabular}{lccc|ccc}
        \toprule
        \multirow{2}*{\bfseries Methods} &
        \multicolumn{3}{c}{MSCOCO} & \multicolumn{3}{c}{NUS-WIDE}\\ 
        \cmidrule{2-7} &16bit&32bit&64bit&16bit&32bit&64bit \\
        \midrule
        HashNet &0.745& 0.773& 0.788 &0.757&0.775&0.790\\
        DCH &0.759 &0.801 &0.825 &0.773 &0.795& 0.818\\
        CSQ &\underline{0.796} &\underline{0.838} &\underline{0.861} &\underline{0.810} &\underline{0.825} &\underline{0.839} \\
        \textbf{COOKIE} &\textbf{0.811}&\textbf{0.884}&\textbf{0.910}&\textbf{0.822}&\textbf{0.852}&\textbf{0.855}\\
        \midrule

    \end{tabular}
   \label{tab:imageretrieval}
\end{table}

\begin{table}[t]\setlength{\tabcolsep}{6pt}
    \fontsize{9.5}{10}\selectfont
    \centering
    \caption{Results of different Pre-training Tasks.}
        \begin{tabular}[t]{ccc|ccc}
              \toprule
              $\mathcal{CCL}$ & $\mathcal{VCL}$ & $\mathcal{TCL}$ & ITM & IR & TM \\
            \midrule
             & & &526.9&0.861&46.1 \\
             \hline
             \checkmark &  & &545.4&0.898&72.4  \\
             \checkmark & \checkmark&  &547.8&\underline{0.909}&72.1  \\
              \checkmark & &\checkmark &\underline{548.1} &0.899&\underline{75.1} \\
             \checkmark & \checkmark & \checkmark & \textbf{550.0} & \textbf{0.910} & \textbf{75.3}\\
             \midrule
        \end{tabular}
   \label{tab:tasks}
\end{table}

\begin{table}[t]\setlength{\tabcolsep}{5pt}
    \fontsize{9.5}{10}\selectfont
    \centering
    \caption{Results whether using Weight-sharing \& TAV TE.}
    \begin{tabular}[t]{l|c}
    \toprule
                  model & ITM \\ 
               \midrule 
                 baseline & 536.6 \\
                 FC Layers (w/ weight sharing)  & 536.8 \\
                 \midrule
                 WS-TE (w/o weight sharing) & 541.6\\
                   WS-TE (w/ weight sharing)& \underline{547.5} \\
                  WS-TE (w/o weight sharing)+TAV-TE & 543.8\\
                 WS-TE (w/ weight sharing)+TAV-TE & \textbf{550.0}\\
                \midrule
    \end{tabular}    
   \label{tab:wste}
\end{table}

\begin{table}[!t]\setlength{\tabcolsep}{10pt}
    \fontsize{9.5}{10}\selectfont
    \centering
    \caption{Ablation study on the Num of Layers of WS-TE.}
    \begin{tabular}[t]{c|c||c|c}
    \toprule
             \multicolumn{2}{c}{\bfseries w/ pre-training}&\multicolumn{2}{c}{\bfseries w/o pre-training}\\  
            \midrule
              num   & ITM & num & ITM\\ 
              \hline
            $0\times$  & 536.6 &$0\times$&\underline{528.7}\\

            $1\times$  & \underline{547.8}&$1\times$&\textbf{534.6} \\
            $2\times$  & \textbf{550.0}&$2\times$&526.9 \\
             $4\times$  & 539.8&$4\times$&510.4 \\
             $8\times$  & 500.7 &$8\times$&250.8\\
             \midrule
    \end{tabular}    
   \label{tab:layers}
\end{table}

\begin{table}[t]\setlength{\tabcolsep}{2pt}
    \fontsize{9.5}{10}\selectfont
    \centering
    \caption{Results with different pre-training corpus.}
    \begin{tabular}[t]{l|ccc}
    \toprule
              dataset & ITM & IR & TM \\ 
            \midrule 
            MSCOCO (11w) & 530.4 &0.895&71.9 \\
             Flickr30K (3w) & 531.7&0.886&69.5 \\
             CC (2.8M) & 540.5 &0.902&74.4 \\
             CC+SBU (3.6M) & 544.3&0.905&\textbf{75.3} \\
             CC+SBU+COCO+F30K (4.2M) & \underline{547.6}&\textbf{0.910}&\underline{75.1} \\
             CC+SBU+COCO+F30K+VQA+GQA (5.9M) & \textbf{550.0}&\underline{0.908}&74.9 \\
             \midrule
        \end{tabular}    
   
   \label{tab:pretrain corpus}
\end{table}

\begin{table}[!h]\setlength{\tabcolsep}{3pt}
    \fontsize{9.5}{10}\selectfont
    \centering
    \caption{Ablation study on Visual Backbones.}
    
    \begin{tabular}[t]{l|ccc||ccc}
            \toprule
            &\multicolumn{3}{c}{\bfseries w/ pre-training}&\multicolumn{3}{c}{\bfseries w/o pre-training}\\  
            \midrule
                  model & ITM & IR & TM & ITM & IR & TM\\ 
                 \midrule
                 ResNet50 & 526.4 & 0.910 & 74.8& 512.8&0.861&46.1\\

                 ResNet101 & \underline{531.3} & \underline{0.911} & \textbf{75.3} & \underline{516.5}&\underline{0.872}&46.1 \\ 
                 ResNeXt101 & \textbf{550.0} & \textbf{0.932} & \underline{75.0} & \textbf{526.9}&\textbf{0.911}&46.1 \\
                 \midrule
            \end{tabular}    
   
   \label{tab:visual backbone}
\end{table}

\paragraph{Text Matching}
We choose Text Matching (TM) task to verify the capability of our pre-trained text encoder. The experiments focus on classic semantic text similarity (STS) \cite{sts} task, which evaluate the effects of textual representation learning by calculating the similarity of two sentences. As illustrated in \cite{sbert}, directly computing the cosine similarity of two text representations is much more efficient than obtaining the similarity score by processing the concatenation of two sentences as did in \cite{bert}. After obtaining two sentence embeddings $\vec{T}_1$ and $\vec{T}_2$  by using the pre-trained models, we directly compute their similarity by cosine similarity. The similarity labels of STS are decimals in range 0-5. We conduct experiments on the widely-used STS12-16 and STS benchmark datasets \cite{sts} to verify the performance. The experimental results are evaluated by the mean values of Pearson and Spearman metrics. The experimental results are shown in Table~\ref{tab:sts}. We compare the proposed method with methods that are only pre-trained on texts such as BERT \cite{bert}, Roberta \cite{roberta}, CLEAR \cite{clear} and approaches such as MACD \cite{macd} which are pre-trained by multi-modal data.


   

\paragraph{Image Retrieval}
Content-based image retrieval (IR) has significant practical application, which is also a hot research direction \cite{dubey2020decade}. To mostly utilize the knowledge learned from our pre-training, we choose two widely-used datasets i.e. MSCOCO \cite{mscoco} and NUS-WIDE \cite{nuswide} to evaluate the performance of image retrieval. Content-based image retrieval requires more understanding of the semantic meanings of the entire picture, especially in fine-grained visual semantic details. We adopt the current sota method CSQ \cite{csq} as our baseline. Unlike CSQ, We substitute the image encoder with our pre-trained image encoder which contains much more information learned from cross-modal data. For a fair comparison, ResNet50 \cite{resnet} backbone is used and images are resized to 224x224. $MAP@5000$ is recorded to evaluate the performance. We compare COOKIE with several state-of-the-art methods including  HashNet \cite{hashnet}, DCH \cite{dch}, and CSQ. 
The experimental results are reported in Table \ref{tab:imageretrieval}.

\paragraph{Performance Comparison with SoTA}
The proposed contrastive cross-modal knowledge sharing pre-training strategy learns universal multi-modal representations for multiple downstream matching and retrieval tasks. To be specific, for cross-modal retrieval, COOKIE obtains new SOTA results on Flickr30K dataset and MSRVTT dataset and achieves competitive results on MSCOCO dataset against Oscar \cite{oscar} with just 3/1000 time of inference. For image-text matching task, when compared to traditional double-stream methods including GPO \cite{gpo} with ResNeXt101 \cite{resnext}, our pre-training paradigm observably improves performances, as shown in Table \ref{tab:coco} and Table \ref{tab:f30k}. Compared to two-stage pre-training methods whose visual features are extracted by Faster R-CNN \cite{faster} such as Uniter \cite{uniter} and Oscar \cite{oscar}, the proposed COOKIE not only has the advantage of speed but also uses less pre-training data (5.9M vs 6.5M\&9.6M). The proposed method also outperforms ResNeXt152-based Pixel-BERT \cite{pixel-bert}. As shown in Table \ref{tab:msrvtt}, The proposed image-text pre-training strategy significantly promotes video-text matching scores on MSRVTT dataset, which improves $R@1$ score of Video-to-text retrieval from $16.0$ to $20.0$ and $9.2$ to $9.8$ for Text-to-Video. As all of these methods use the same visual feature, the proposed method significantly improve the accuracy of video-text matching.  For those methods with stronger backbone exacting visual features (MEE \cite{mee}, CE \cite{ce}, and W2VV$_{++}$ \cite{w2vv}), our COOKIE also exceeds them in some metrics, especially for the Video-to-Text retrieval task.

Furthermore, COOKIE also obtains new SOTA results for unimodal matching and retrieval tasks, demonstrating the effect of the proposed cross-modal knowledge sharing strategy. For the text matching task, the method achieves a performance gain of 3.9\% on STS-B and more pronounced improvements on five datasets from STS12 to STS16, as shown in Table \ref{tab:sts}. It should be noted that these methods (BERT, RoBERTa, and CLEAR) only use text corpora for training. Experimental results show that the proposed cross-modal pre-training strategy successfully achieves visual-semantic sharing with the text encoder. COOKIE also outperforms MACD\cite{macd} using a similar cross-modal pretraining strategy. Meanwhile, for the content-based image retrieval task, we obtain 5.7\% and 1.9\% improvement on MSCOCO dataset and NUSWIDE dataset with 64-bit features, respectively. The results are shown in the table \ref{tab:imageretrieval}. All of these performance improvements are achieved by contrastive cross-modal knowledge sharing.

\subsection{Ablation Study}
To explore the performance of COOKIE, we do several ablation studies under various model settings. For the pre-training stage, the default backbones are ResNeXt101 \cite{resnext}, BERT-base \cite{bert}, together with the TAV-TE and the 2-layer WS-TE. We use the full three contrastive losses and the full pre-training dataset to train the model. For image-text matching (ITM), the initial setting of the network is the same as that of pre-training. For image retrieval (IR) and text matching (TM), we substituted the visual backbone with ResNet50 and ResNet101 respectively. Since 
video-text matching is similar to ITM, we choose ITM to represent the results of cross-modal retrieval.


\paragraph{Effectiveness of Three Contrastive Losses}
During pre-training, we design three kinds of contrastive losses to supervise the training process. Cross-modal contrastive learning (CCL) is designed for developing a common subspace and bridging the heterogeneous gap between the two modalities, while visual and textual contrastive learning (VCL \& TCL) aim to keep the knowledge that the uni-modal encoders originally learned from the respective modality. As shown in Table \ref{tab:tasks}, unimodal tasks IR and TM rely on both cross-modal and single-modal contrastive learning, while ITM benefits more from CCL. 

\paragraph{Transformer Encoders}
We apply two special kinds of transformers in the model. As seen in Table \ref{tab:wste}, these two parts are both helpful for cross-modal alignment. Firstly the TAV-TE transfers visual features' distribution. Then 
WS-TE makes sure that the image and text pay attention to the same semantic information. Noted that simple weight-shared fully-connected layers bring no improvement, which proves that the combination of transformer encoder and weight sharing is the essence.


\paragraph{Depth of WS-TE}
The depth of the transformer encoder is a key point for its performance. Here we try to explore the optimal number of layers for WS-TE, as shown in Table \ref{tab:layers}. For models without pre-training, the best setting is only $1$ layer. With the help of pre-training, the network goes deeper, resulting in the best depth of $2$ layers. In the case WS-TE is trained from scratch, the magnitude of the pre-training corpus determines the maximum depth of WS-TE. According to the number of image-sentence pairs in this work, the depth of two layers should be a bottleneck.

\paragraph{Size of Pre-training Corpus}
For cross-modal pre-training, the number of image-sentence pairs plays a crucial role \cite{oscar}. Previously, Uniter \cite{uniter} used $9.6M$ pairs and Oscar uses $6.5M$. In Table \ref{tab:pretrain corpus}, We record experimental results on three tasks with the size of the pre-training corpus changed. It is noticed that the improvement brought by VQA and GQA datasets is tiny. The reason is images of these datasets are coupled with questions instead of captions. The semantics are not quite correlated. Besides, the growth of the data scale can't bring salient performance improvement in single-modal tasks. At the same time, ITM pretrained with Flickr30K(531.7) outperforms that with MSCOCO(530.4). This is because we do experiments on MSCOCO test set and pre-training on Flickr30K provides extra out-domain data.


\begin{figure}[t]
\begin{center}
   \includegraphics[width=1.0\linewidth]{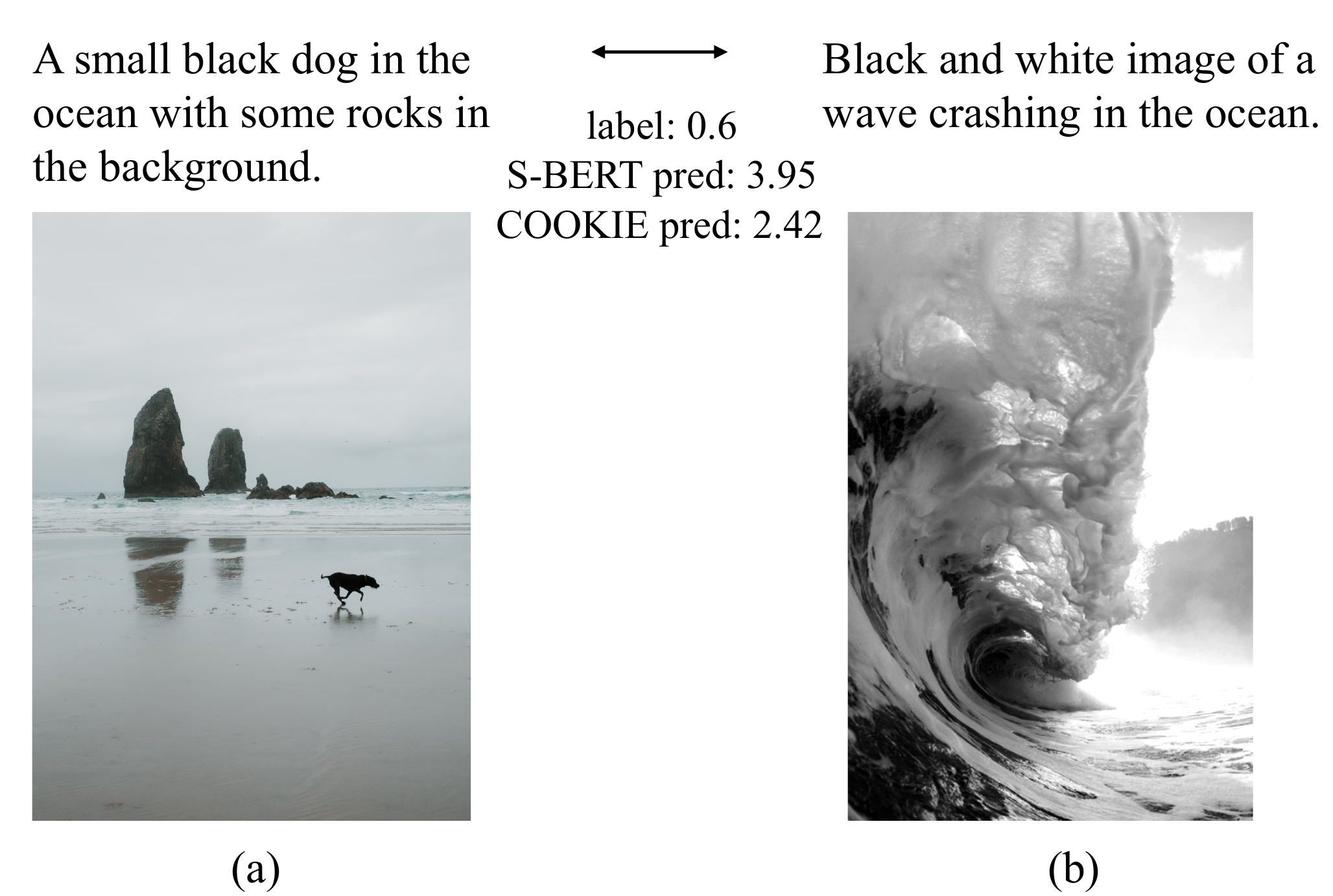}
\end{center}
   \caption{Another example of the proposed knowledge sharing. The two captions have some common words like ``ocean'' and ``black'', however, the semantic information them are quite different. With just languages, it's difficult to judge such dissimilarity. By observing the corresponding images, it's more intuitionistic to distinguish the difference between two descriptions. The similarity score ranges from 0 to 5.}
\label{fig:intro-2}
\end{figure}

\paragraph{Different Visual Backbones}
In order to validate the robustness of our method, we alter the visual backbones. In Table \ref{tab:visual backbone}, as expected, COOKIE performs better on ITM and IR as the visual backbone gets stronger. However, for TM, stronger visual backbones don't bring performance improvement. We infer the reason is the textual encoder may lose too much original information if the visual backbone is too powerful. For the language part, limited by insufficient computing resources, we only use the BERT-base model. In the future, we will conduct pre-training with the BERT-Large model.

\begin{figure}[t]
\centering
\includegraphics[width=1\linewidth]{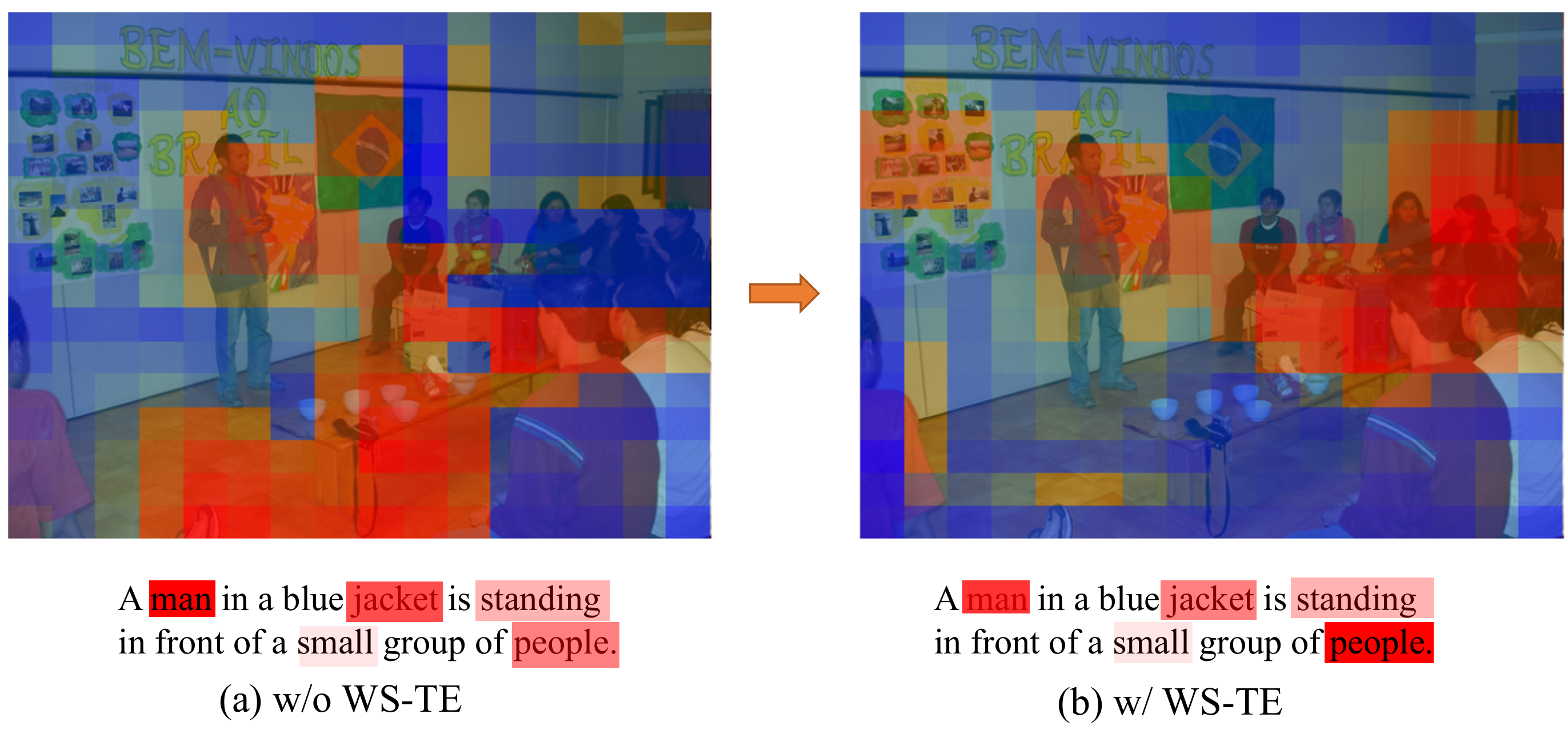}
\caption{Illustration of weight-sharing transformer encoder's effect. Ares paid more attention are marked brighter. As seen, WS-TE forces vision and language concentrate on the same semantics.}\label{ws-te}
\end{figure}

\begin{figure}[t]

   \includegraphics[width=1.0\linewidth]{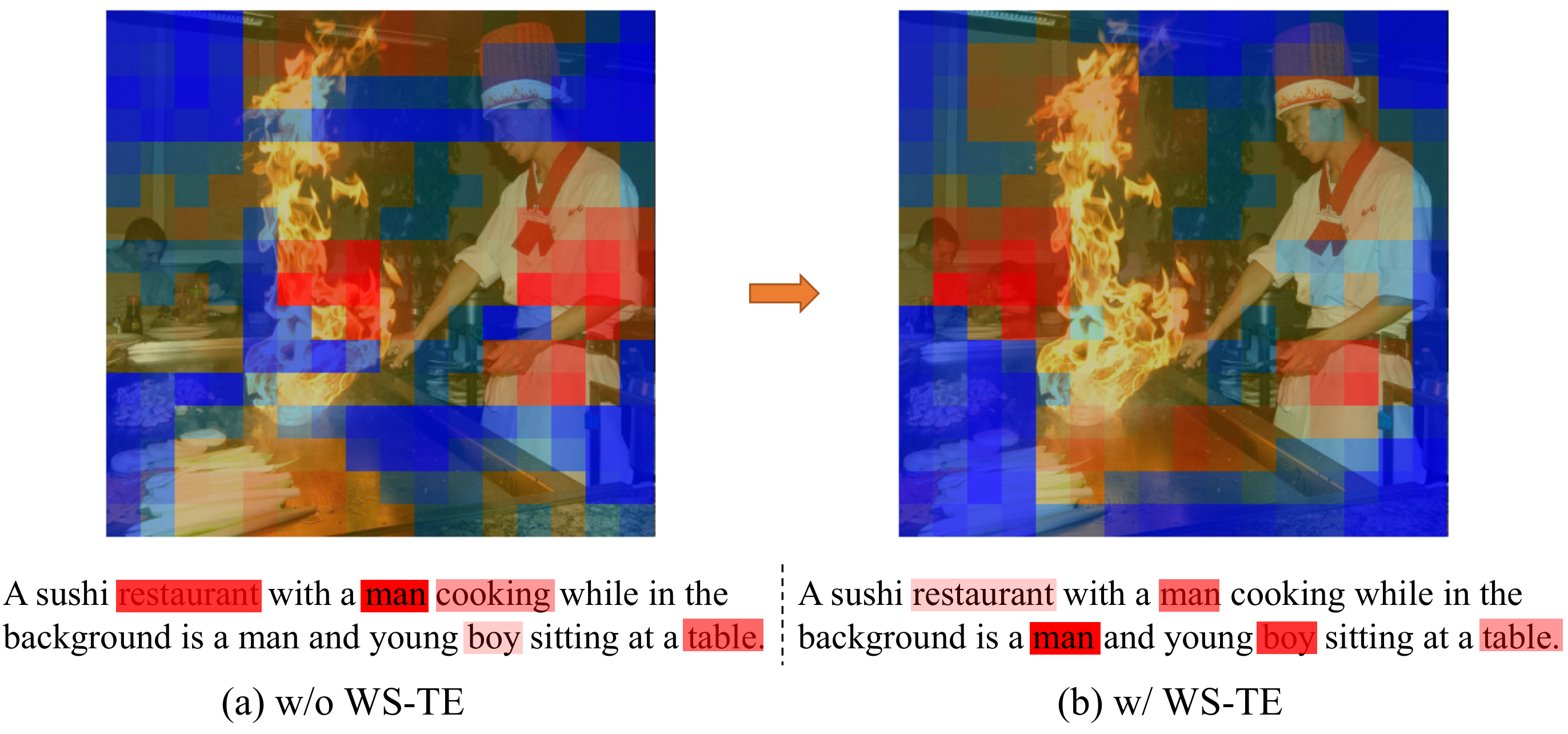}

   \caption{Another illustration of effect of WS-TE. The image should concentrate more on the cook and two people behind him instead of the fire or the table to align with the description.}
\label{fig:ws-te-2}
\end{figure}

\begin{figure}[t]
\centering
\includegraphics[width=1\linewidth]{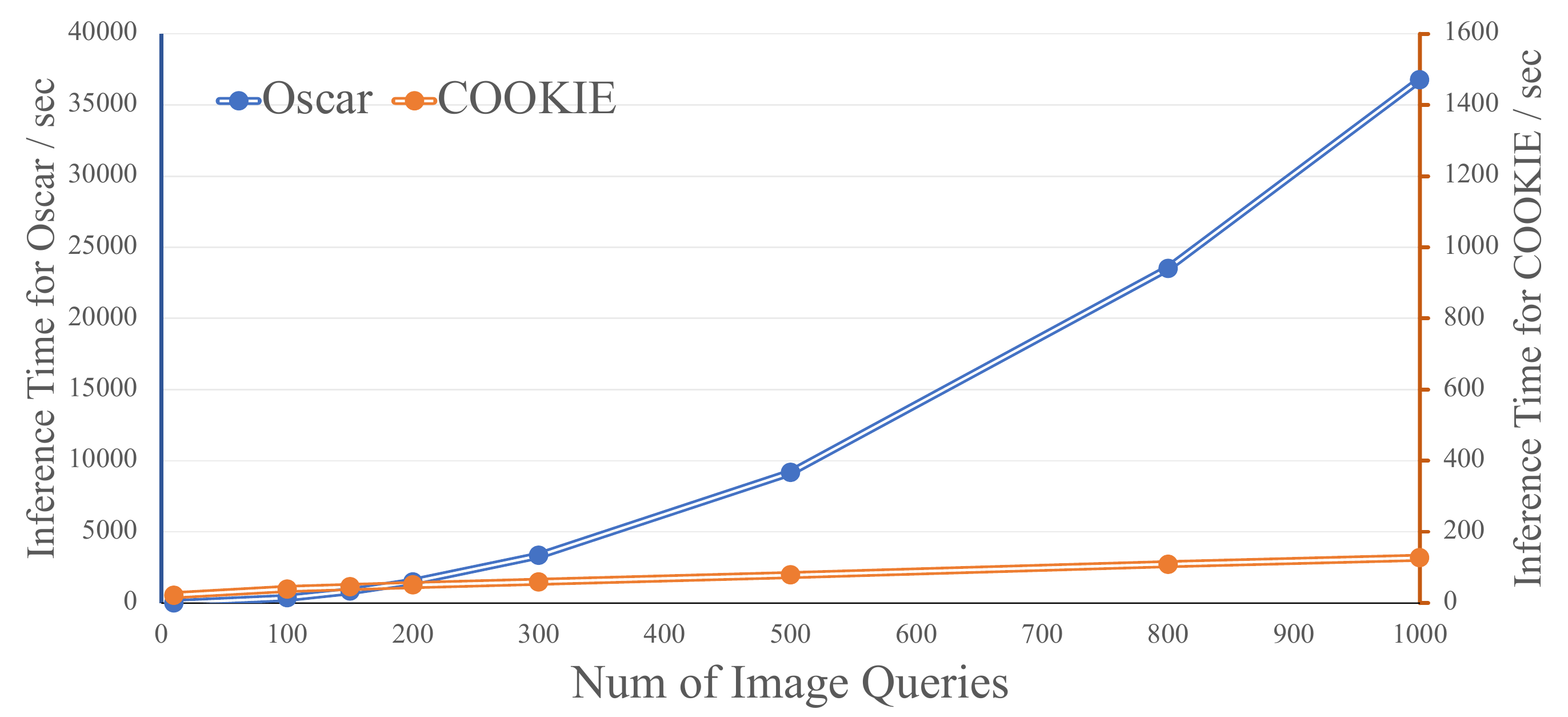}
\caption{Comparison of inference time against Oscar. Experiments are conducted on Flickr30K test set.}
\label{time}
\end{figure}

\subsection{Analysis}\label{ws-te-ana}

\paragraph{Analysis of Knowledge Sharing}
As seen in Fig.~\ref{fig:intro-2}, the two sentences have different semantic meanings although they have many shared words like ``ocean'' and ``black''. The single-modal method Sentence-BERT \cite{sbert} easily judges from these words and gives a similarity score of 3.95, but the label just got 0.6. Judging through mere texts is difficult. With the information provided by images, it could tell that (a) pays attention to ``rock'' and ``dog'' while (b) places more emphasis on ``ocean'' and ``wave''. Equipped with the cross-modal contrastive learning, the similarity value given by COOKIE descends to 2.42. And the result is less than half of the total score of 5.


\paragraph{Analysis of Weight-Sharing Transformer}
A weight-sharing transformer encoder (WS-TE) is devised on the top of the whole network. Though there is heterogeneous cross-modal gap between texts and images, the weight-sharing attention mechanism leads the two paths focusing on the tokens depicting similar semantics. In Figure ~\ref{ws-te}, the attention yielded by the WS-TE is visualized. 
Following \cite{vsrn,dsran}, considering the final representation contain more information of the salient parts of the texts or images, the similarities of the yielded representation $\vec{I}$ or $\vec{T}$ between the tokens output by the WS-TE are computed. Now each area or word has a similarity score with the final feature. 
Then we rank the areas by their scores. In the figures, the areas with higher ranks are marked by brighter color. The top-5 ranked words of sentences are marked. As the figure on left shown, the sentence and image attend to different semantics without WS-TE. 
In the sentence, ``people'' ``jacket'' and ``man'' are primary words containing semantics. However, in the image, irrelevant ``flag'' and ``table'' are given more attention. As the figure on right shown, images and texts are prone to pay attention to the same semantics, i.e., ``man'' and ``group of people'' with the WS-TE. 

Another visualization of the effect of the weight-sharing transformer encoder (WS-TE) can be seen in Fig.~\ref{fig:ws-te-2}. The transformer encoders on each path help the image and text attend to different semantics without weight sharing. With weight sharing, tokens which contain similar semantics are assigned similar attention weights, thus the two modalities similarly attend to the cook and two people in the background, which are easy to be neglected.

\paragraph{Analysis of Inference Time}

COOKIE is a double-tower method without the calculation of cross-modal interaction. So it greatly speeds up the inference of cross-modal retrieval. And we do experiments on the different proportion of Flickr30K test set to compare the inference time (similarity computing plus feature extraction). As seen in Fig.~\ref{time}, the one-stream method (Oscar \cite{oscar}) have an $O(n^2)$ time complexity compared to $O(n)$ of our COOKIE.


\section{Conclusion}
This paper proposes a novel Crontrastive Cross-Modal Knowledge Sharing Pre-training (COOKIE) method. The goal is to learn universal separate vision and language representations for downstream multi-modal retrieval tasks. Here we focus on cross-modal retrieval, image retrieval, and text matching. Firstly, to better align the semantics of vision and language modalities, we design a TAV-TE and a weight-sharing transformer encoder. And at the same time, we pre-train the model with cross-modal contrastive learning and single-modal contrastive learning using publicly available image-sentence pairs with a magnitude of 5.9M. Our COOKIE reaches comparable results on cross-modal retrieval with only 3/1000 inference time and simultaneously achieves new state-of-the-art results on uni-modal matching tasks text matching and image retrieval.

In the future, we plan to apply our pre-training to more multi-modal tasks such as visual question answering and image captioning. We also plan to research more within-modal basic understanding tasks such as image classification and classic GLUE tasks.




\section*{Acknowledgment}

This work is supported by National Natural Science Foundation of China under grants 61771145.

\ifCLASSOPTIONcaptionsoff
  \newpage
\fi



\bibliographystyle{IEEEtran}
\end{document}